\documentclass[lettersize,journal]{IEEEtran}
\usepackage{amsmath,amsfonts}
\usepackage{algorithmic}
\usepackage{algorithm}
\usepackage{array}
\usepackage[caption=false,font=normalsize,labelfont=sf,textfont=sf]{subfig}
\usepackage{textcomp}
\usepackage{stfloats}
\usepackage{url}
\usepackage{verbatim}
\usepackage{graphicx}
\usepackage{cite}
\usepackage{booktabs}
\usepackage{amsfonts}
\usepackage{graphicx}
\usepackage{amsmath}
\usepackage{arydshln}
\usepackage{multirow}
\graphicspath{{figures/}}
\usepackage{bbding}
\usepackage[usenames,dvipsnames]{color}
\usepackage{colortbl}
\definecolor{mygray}{gray}{.9}

\begin{document}

\title{Class Incremental Learning with Self-Supervised Pre-Training and Prototype Learning}

\author{Wenzhuo Liu,
	Xinjian Wu,
	Fei Zhu,
	Mingming Yu,
	Chuang Wang,
	Cheng-Lin Liu~\IEEEmembership{Fellow,~IEEE}
	\IEEEcompsocitemizethanks{\IEEEcompsocthanksitem The authors are with the University of Chinese Academy of Sciences, Beijing, P.R. China, and the State Key Laboratory of Multimodal Artificial Intelligence Systems, Institute of Automation, Chinese Academy of Sciences, 95 Zhongguancun East Road, Beijing 100190, P.R. China. \protect \\
		Email: liuwenzhuo20@mails.ucas.ac.cn, \{wuxinjian2020, zhufei2018, yumingming2020\}@ia.ac.cn \{chuangwang, liucl\}@nlpr.ia.ac.cn}
}

%


\maketitle

\begin{abstract}
Deep Neural Network (DNN) has achieved great success on datasets of closed class set. However, new classes, like new categories of social media topics, are continuously added to the real world, making it necessary to incrementally learn. This is hard for DNN because it tends to focus on fitting to new classes while ignoring old classes, a phenomenon known as catastrophic forgetting. State-of-the-art methods rely on knowledge distillation and data replay techniques but still have limitations.
In this work, we analyze the causes of catastrophic forgetting in class incremental learning, which owes to three factors: representation drift, representation confusion, and classifier distortion. Based on this view, we propose a two-stage learning framework with a fixed encoder and an incrementally updated prototype classifier.
The encoder is trained with self-supervised learning to generate a feature space with high intrinsic dimensionality, thus improving its transferability and generality. The classifier incrementally learns new prototypes while retaining the prototypes of previously learned data, which is crucial in preserving the decision boundary.Our method does not rely on preserved samples of old classes, is thus a non-exemplar based CIL method.
Experiments on public datasets show that our method can significantly outperform state-of-the-art exemplar-based methods when they reserved 5 examplers per class, under the incremental setting of 10 phases, by 18.24\% on CIFAR-100 and 9.37\% on ImageNet100. \textit{This paper has been under review by a journal since 19-Apr-2023.}
\end{abstract}



\section{Introduction}
\label{sec:introduction}
\IEEEPARstart{D}{eep} Neural Network (DNN) plays a vital role in various pattern recognition tasks, such as image classification\cite{krizhevsky2012imagenet,he2016deep}, objection detection\cite{ren2015faster, redmon2016you}, and instance segmentation\cite{long2015fully,he2017mask}. 
In the paradigm of end-to-end training, the parameters of the feature encoder and classifier are optimized jointly from raw data. 
This kind of joint learning leads to superior classification performance on datasets with fixed class set where the distribution of training data matches the sample distribution at the test phase. 

However, DNNs' performance deteriorates in real-world scenarios where data is often in a stream format from open environments or temporarily available for privacy reasons\cite{delange2021continual,9724647,9181489}. 
For example, social media platforms are constantly witnessing the emergence of new hot topics, while online shopping environments are continually evolving with changes in consumer behavior and preferences.
In such cases, models should be able to learn incrementally, rather than relying on pre-compiled data and retraining.

Class incremental learning (CIL), also called continual learning, in which new-class data are learned sequentially and old-class data can not be reused or only a small fraction of old-class data can be reused, poses a major challenge called, \textit{catastrophic forgetting}\cite{8963851,9756660,parisi2019continual,french1999catastrophic,robins1995catastrophic,mccloskey1989catastrophic,zhangCvpr22ContinuSSeg}, {\em i.e.} the classification accuracy of the old classes dramatically decreases alongside the training for new classes.
\begin{figure}[t]
  \centering
  \includegraphics[width=1.0\linewidth]{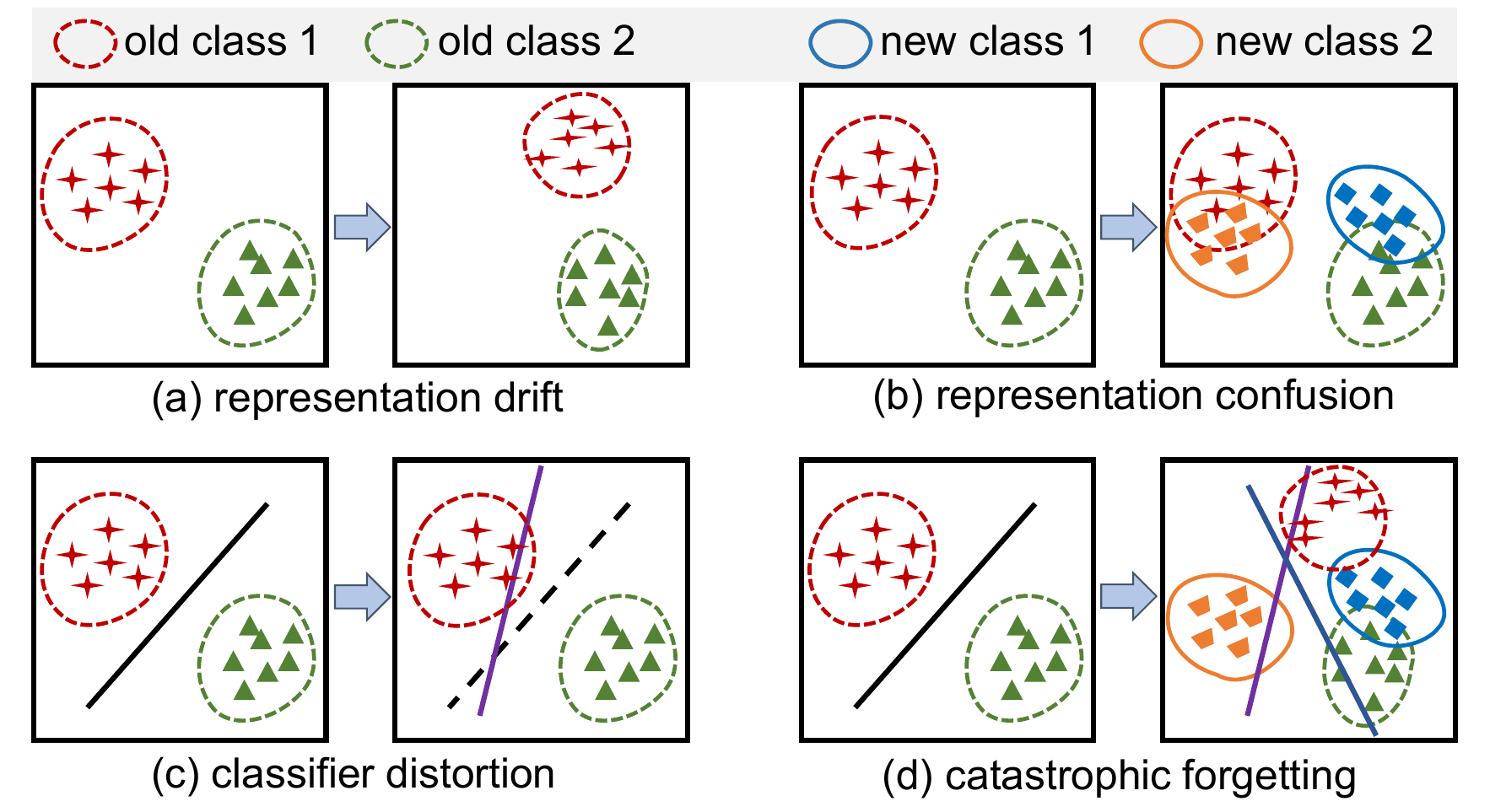}
  \vskip -0.1in
  \caption{In the process of CIL, the model parameters are updated continuously, which leads to three phenomenons: (a) representation drift, (b) representation confusion, and (c) classifier distortion. Then there will be (d) catastrophic forgetting problem.}
  \vskip -0.1in
  \label{fig:1}
\end{figure}
Previous works tried to alleviate catastrophic forgetting mainly via three approaches: parameter based\cite{kirkpatrick2017overcoming,aljundi2018memory,li2017learning,dhar2019learning}, data replay based\cite{belouadah2018deesil,belouadah2019il2m,rebuffi2017icarl,castro2018end,9145832} and architecture based\cite{rusu2016progressive, mallya2018packnet, serra2018overcoming, yoon2017lifelong}. State-of-the-art methods\cite{rebuffi2017icarl,hou2019learning,douillard2020podnet} mostly adopt two strategies: \textit{knowledge distillation} (KD)\cite{hinton2015distilling} and \textit{data replay}.
Whereas, KD requires that the new model is not much different from the old model, it focuses on capturing features favorable for old classes and may hurt the accuracy on new classes\cite{ahn2020ss}.
On the other hand, storing data in replay-based (or called exemplar-based) methods is undesirable due to memory limits or privacy issues, which prevents the old-class data to be stored\cite{zhu2021prototype,10058177}.

By analyzing the representation and classifier learning in CIL, the phenomenon of catastrophic forgetting can be attributed to three main factors, as illustrated in Figure \ref{fig:1}.
 (1) \textbf{Representation drift}, which means that the varying feature extractor causes the change of representation of old classes; (2) \textbf{representation confusion}, of which the samples of new classes and those of old classes are overlapped in the feature space; and (3) \textbf{classifier distortion}, which indicates that the incremental update of the classifier on new-class samples distort the weights of old classes. The combined effect of the above three factors leads to catastrophic forgetting.

To address the above three key issues in CIL, we propose a two-stage learning framework that consists of a pre-trained feature extractor and an incrementally updated prototype classifier. Our method effectively tackles these issues as follows:

\textbf{First}, our approach addresses representation drift by leveraging a pre-trained feature extractor obtained through self-supervised learning (SSL) and keeping the feature extractor fixed throughout CIL.
\textbf{Second}, the feature space acquired by SSL enables effective representation and separation of both old and new classes, thereby mitigating representation confusion. As analyzed in Section \ref{subsec: Learning representations via SSL}, SSL yields a more comprehensive and class-agnostic feature space, featured by its higher intrinsic dimensionality and reduced similarity between distinct classes.
\textbf{Third}, to alleviate classifier distortion in CIL, we design the Incremental Prototype Classifier (IPC), which balances discriminative and generative aspects during learning and employs an incremental update strategy for continual adaptation, maintaining appropriate decision boundaries in the CIL process.

Without preserving data of old classes, our prototype-based CIL method is non-exemplar based, but can even outperform exemplar-based methods. Experimental results on public image datasets demonstrate that our proposed method achieves significant performance improvement \textbf{\textit{without retaining samples from old classes or utilizing the old model for KD}}. It is simple, effective, and compatible with most self-supervised models, surpassing state-of-the-art methods by a considerable margin in fair comparison. We summarize the main contributions as follows.

\begin{enumerate}
\item[$\bullet$] We introduce a two-stage framework for CIL with a pre-trained feature encoder which yields high intrinsic dimensionality and enables separation of unseen classes.
\item[$\bullet$] We present an Incremental Prototype Classifier (IPC) to address classifier distortion within this framework, and analyze why it is suitable for CIL.
\item[$\bullet$] The experimental results demonstrate that our \textbf{non-exemplar based} method fairly outperforms state-of-the-art CIL methods, regardless of whether they preserve old-class data or not.
\end{enumerate}

The rest of this paper is organized as follows. We provide an overview of the related work in Section \ref{sec: 2} and define the CIL problem in Section \ref{sec: 3}. Section \ref{sec: 4} describes the proposed approach. Section \ref{sec: 5} presents our experimental results, and in Section \ref{sec: 6} draws concluding remarks.

\section{Related Work}
\label{sec: 2}

\subsection{Class Incremental Learning}
In class incremental learning (CIL), the model learns incrementally from the data of new classes while preserving the knowledge of old classes. Since no data or little data of old classes is reserved, the ability of old classes will deteriorate. This raises the so-called problem of catastrophic forgetting problem \cite{9705128,french1999catastrophic,mccloskey1989catastrophic,robins1995catastrophic}. Many CIL methods have been proposed to handle the catastrophic problem. The recent reviews can be found in\cite{zhu2023}. Existing works can be divided into three branches: parameter based, data replay based, and architecture based, considering whether to use the data of the old category and whether to dynamically adjust the network in the CIL process, respectively.

\textbf{\textit{Parameter-based methods}} regularize network parameters either explicitly \cite{kirkpatrick2017overcoming,zenke2017continual,aljundi2018memory} or implicitly \cite{li2017learning,dhar2019learning,liu2020more}. Explicit methods aim to prevent drastic changes to important parameters in the old model while learning new classes. Implicit methods \cite{li2017learning,dhar2019learning,liu2020more}, on the other hand, are based on knowledge distillation \cite{hinton2015distilling}.

\textbf{\textit{Data replay-based methods}}, also called as exemplar based methods,  use a small portion of old data to jointly train the model with new data. Some works \cite{belouadah2018deesil,belouadah2019il2m} directly use these data for subsequent training. More works \cite{rebuffi2017icarl,castro2018end,douillard2020podnet,hou2019learning,liu2020mnemonics,simon2021learning,hu2021distilling} introduce a distillation loss to preserve the old knowledge of the model. An alternative solution that “stores” old knowledge in generative models \cite{kamra2017deep,shin2017continual} is synthesizing samples. However, storing real data is not feasible due to limited storage resources, and training big generative models for complex datasets is inefficient. Some methods do not reserve real data of old classes but reserve or synthesize feature data of old classes\cite{zhu2023}.

\textbf{\textit{Architecture-based methods}} dynamically adjust the network structure during CIL \cite{rusu2016progressive, mallya2018packnet, serra2018overcoming, yoon2017lifelong}. However, changing the architecture during training is often infeasible for a large number of tasks, especially in CIL. In this paper, we aim to build an incremental learner without storing old data, leveraging complex generative models, or using old models for knowledge distillation.

Some recent works, such as L2P\cite{wang2022learning}, DualPrompt\cite{wang2022dualprompt}, CODA-Prompt\cite{smith2022coda}, and S-iPrompts\cite{wang2022s}, have employed pre-trained models in CIL, significantly improving accuracy. However, these works have insufficiencies like the absence of class exclusion from pre-trained datasets during incremental learning, inconsistency in backbone usage compared to other continual learning methods, and a lack of comparison with pre-trained models for contrastive learning methods, resulting in unfair comparisons.

\subsection{Self-supervised Learning}
Self-supervised learning (SSL) has been widely used for learning representations from unlabeled data. The process starts by training general representations through a self-supervised task, such as predicting rotations \cite{gidaris2018unsupervised}, relative patch positions \cite{doersch2015unsupervised}, image colorization \cite{larsson2016learning}, or inpainting \cite{pathak2016context}. These representations are then fine-tuned using minimal labeled data for a downstream supervised task.

Recent SSL methods can be divided into four categories: contrastive learning, negative-free methods, clustering-based methods, and redundancy reduction-based methods. 

\textbf{\textit{Contrastive learning methods}}, including CPC\cite{oord2018representation}, AMDIM\cite{bachman2019learning}, CMC\cite{tian2020contrastive}, SimCLR\cite{chen2020simple}, and MoCo\cite{he2020momentum}, generate both positive and negative pairs from augmented views of the input data. The model is trained to output similar representations for positive pairs and different representations for negative pairs. However, this requires large batch sizes or memory banks to generate enough negative pairs to prevent representational collapse.

\textbf{\textit{Negative-free methods}}, notably BYOL \cite{grill2020bootstrap} and SimSiam \cite{chen2021exploring}, can learn robust representations through the use of only positive pairs, eliminating the need for negative pairs.
	These methods employ a dual pair of Siamses networds: the representation of two views are trained to match, one obtained by the composition of an online and predictor network, and the other by a target network. The target network is not trained via gradient descent; and either employs a direct copy of the online network, or a momentum encoder that slowly follows the online network in a delayed fashion through an exponential moving average. Compared to contrastive learning, they are generally more efﬁcient and conceptually simple while maintaining state-of-the-art performance\cite{tian2021understanding}.

\textbf{\textit{Clustering-based methods}}, such as SwAV \cite{caron2020unsupervised}, DeepCluster v2 \cite{caron2018deep}, and DINO \cite{caron2021emerging}, utilize unsupervised clustering algorithms to align representations with their corresponding prototypes. The prototypes are learned by aggregating the feature representations within each cluster. As a result, representations are trained to resemble their prototypes, leading to enhanced generalization and robustness. However, these methods suffer from high computational cost and can be impacted by the initialization of clustering.

\textbf{\textit{Redundancy reduction-based methods}}, like BarlowTwins\cite{zbontar2021barlow} and VicReg\cite{bardes2022vicreg}, reduce the correlation between different representations. BarlowTwins optimizes a cross-correlation matrix to be a unit matrix, making positives similar and negatives orthogonal, while VicReg uses a combination of variance, invariance, and covariance regularizations.

The combination of self-supervised learning with open-world challenges, such as continual learning and out-of-distribution recognition, has received considerable attention. Studies like SSLR\cite{chen2021understanding} have demonstrated the advantages of incorporating self-supervised tasks in enhancing feature spaces and out-of-distribution recognition. The PASS method (Prototype Augmentation and Self-Supervision)\cite{zhu2021prototype} showed the effectiveness of using self-supervised learning in mitigating forgetting. The CaSSLe (Self-Supervised Models are Continual Learners)\cite{fini2022self} and Continual BarlowTwins\cite{marsocci2022continual} approach, which combines knowledge distillation, data replay, and self-supervised learning, enables the self-supervised model to adapt to new knowledge while retaining old knowledge.

In this work, we exploit the task-agnostic features of SSL to mitigate representation drift and confusion in continual learning. To assess its effectiveness, pre-trained models from prior studies were added to the comparison with state-of-the-art continual learning methods.

\subsection{Prototype Learning}
Prototype learning aims to learn a prototype vector to represent each class, via clustering\cite{macqueen1967classification}, Learning vector quantization\cite{kohonen1995learning}, or more discriminative learning methods. These representative values, called prototypes, are used to identify samples or other tasks.\par
A recent prototype learning method, convolutional prototype network (CPN)\cite{yang2020convolutional} applies the idea of discriminative learning quadratic discriminant function (DLQDF) to convolution networks for learning prototypes in deep feature space under the losses of discrimination and class representation.\par
Prototype learning is also widely used in class incremental learning. Many CIL methods, such as iCaRL\cite{rebuffi2017icarl}, LUCIR\cite{hou2019learning}, PODnet\cite{douillard2020podnet}, Coil\cite{zhou2021co}, and DER\cite{yan2021dynamically}, relies on nearest-prototype classifier in classification. State-of-the-art non-exemplar based methods, such as PASS\cite{zhu2021prototype}, SSRE\cite{zhu2022self}, and Fusion\cite{toldo2022bring}, use prototypes to pseudo samples of old classes. However, these methods do not learn prototypes directly under the discriminative loss.\par
Inspired by DLQDF and CPN, our incremental prototype classifier (IPC) is trained in the feature space of self-supervised learning, considering discriminative loss and prototype learning loss concurrently and updating continually. 
IPC is shown to mitigate classifier distortion without preserving old data and is compatible with most of the self-supervised models, to boost the performance of CIL signiﬁcantly.

\begin{figure*}[!t]
  \centering
  \includegraphics[width=1.0\linewidth]{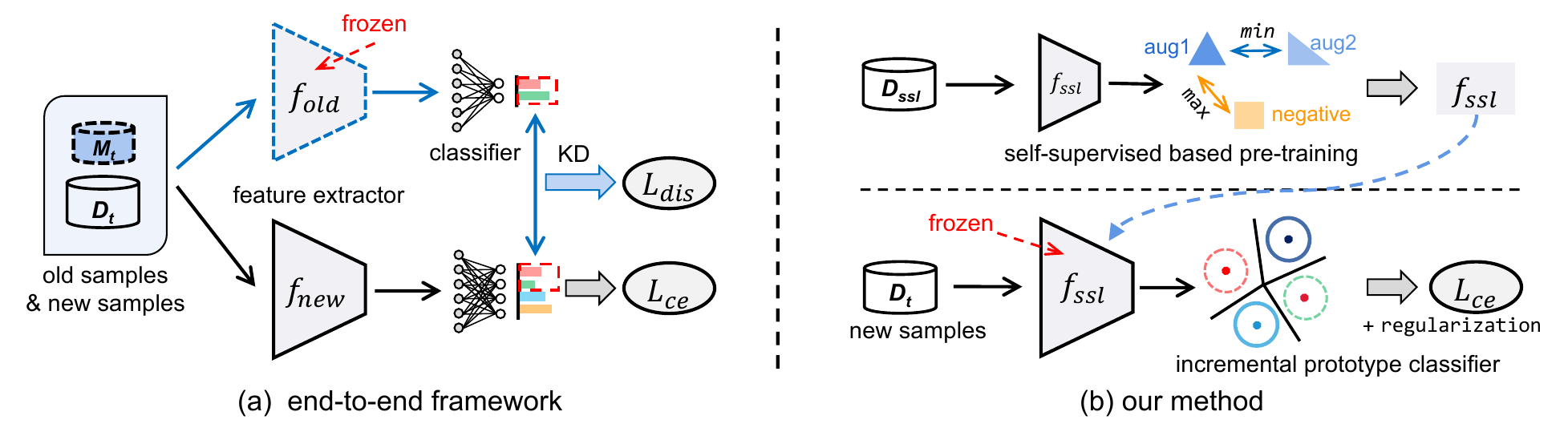}
  \vskip -0.1in
  \caption{Illustration of the end-to-end framework\cite{hou2019learning,rebuffi2017icarl,douillard2020podnet,simon2021learning} and our two-stage framework in CIL. We model the training process of the previous model $\Theta_{t-1}$ to the current model $\Theta_{t}$. In the end-to-end model, the feature extractor and classifier parameters are trained jointly with reserved samples and KD loss. In our framework, the parameters of the feature extractor are frozen, and we maintain old prototypes while learning prototypes of the new categories. Our method is non-exemplar based, simple and effective.}
  \vskip -0.1in
  \label{fig:2}
\end{figure*}

\section{Preliminaries}
\label{sec: 3}
\subsection{Problem Definition}
The CIL problem consists of a sequence of $T$ tasks with disjoint datasets. Each task contains a dataset $\mathcal{D}_t$ with new classes. At phase $t$, the model is updated by the training set in $\mathcal{D}_t$ and is required to classify all the exposed test samples in $\{\mathcal{D}_i\}_{i\leq t}$. Therefore, the number of classes for the classifier increases as the task shifts.

Due to the storage limitation and private issues, usually the model can not use the data in previous tasks. However, some methods such as \cite{hou2019learning, rebuffi2017icarl, douillard2020podnet, hu2021distilling} ease the constraints by preserving a small subset of previous samples for the current task $T$ in $\mathcal{M}_t$, and re-train the model using both the new data and the memory (\textit{i.e.,} $\mathcal{D}_t \cup \mathcal{M}_t$).

To facilitate analysis, we represent the DNN based model with two parts: a feature extractor and a classifier. The feature extractor parameterized by $\theta$ plays the role of encoding the input into a descriptor as $z=f_{\theta}(x)$. Then the descriptor is passed through the classifier parameterized with $\varphi$ to obtain the probability distribution $g_{\varphi}(z)$ for $x$. Denote the overall parameters by $\Theta = (\theta,\varphi)$. At the time $t$, the general goal is to update model $\Theta_t$ from the old $\Theta_{t-1}$.

\subsection{Evaluation metrics}
We use the standard average accuracy metrics to measure the performance of CIL, following the iCaRL\cite{rebuffi2017icarl}. Average accuracy is the average of accuracies of all learned tasks, each for classifying all the classes that have already been learned. Specifically, it is defined as $\bar{A} = \frac{1}{T}\sum_{t=1}^{T}{A}_t$, where $A_t$ is the classification accuracy evaluated on the current test set at the $t$-th step, the test set contains the classes seen from $\mathcal{D}_1$ up to $\mathcal{D}_t$. 

For fair comparison, we propose a procedure for evaluating the use of pre-trained models in class incremental learning (CIL). Firstly, for a given multi-class incremental classification dataset, \textbf{\textit{the pre-trained model should not be trained on any of the relevant categories.}} Secondly, to ensure comparability, the same network structure should be used for all CIL methods being compared. Additionally, all CIL methods should use the same pre-trained model initialization.

\section{Our Approach}
\label{sec: 4}
In this Section, we introduce and analyze the proposed CIL framework.

\subsection{Overview of the framework}
We propose a two-stage learning framework with a fixed feature extractor and an incrementally updated prototype classifier. It is compatible with most existing self-supervised learning methods, e.g., BYOL\cite{grill2020bootstrap}, MoCoV2\cite{he2020momentum}, SimCLR\cite{chen2020simple}, SwAV\cite{caron2020unsupervised}, BarlowTwins\cite{zbontar2021barlow}, and SimSiam\cite{chen2021exploring}. 
  SSL learns task-agnostic and transferable representations that help alleviate representation confusion for CIL tasks. The encoder is fixed after the pre-train stage to prevent representation drift, and the classifier learns a prototype for each class considering robustness and discriminability concurrently and updating continually.
   For inference, classification is decided by the nearest prototype according to the Euclidean distance. 

Figure \ref{fig:2} illustrates the commonly used end-to-end incremental framework and our two-stage framework. The innovation of our framework mainly lines in two aspects. First, different from existing end-to-end incremental learning methods, such as iCaRL\cite{rebuffi2017icarl}, LUCIR\cite{hou2019learning}, PODnet\cite{douillard2020podnet}, and PASS\cite{zhu2021prototype}, the encoder is trained on incremental task-agnostic data by self-supervised learning and is frozen after that. Secondly, classification is performed by incremental prototype classifier (IPC) instead of nearest-mean-of-examples classifier (NME) or cosine linear classifier. Further, our method is non-exemplar-based, simple, and effective.

\subsection{Learning representations via SSL}
\label{subsec: Learning representations via SSL}
SSL has shown success in many recognition tasks  for its ability of learning feature representation from unlabeled data. Our study reveals that SSL improves the CIL task by mitigating representation drift and confusion.
Taking BYOL as an example, it consists of an online network, $f_{\theta}$, and a target network, $f_{\xi}$, with respective parameters $\theta$ and $\xi$. The target network, an exponential moving average of the online network, updates as follows:
\begin{equation}
\xi \leftarrow m\xi + (1 - m)\theta,
\end{equation}
where $m$ is the momentum coefficient. Both networks incorporate a backbone network, $g$, and a projection head, $h$. Given an input image $x$, we generate two augmented views, $x_i$ and $x_j$. The objective is to minimize the distance between predicted latent vector $z_i$ and target latent vector $z_j$:
\begin{equation}
\mathcal{L}(\theta) = \frac{1}{N} \sum_{i=1}^{N} \Big\lVert \frac{z_i}{\lVert z_i \rVert} - \frac{z_j^{\xi}}{\lVert z_j^{\xi} \rVert} \Big\rVert^{2},
\end{equation}
where $N$ is the batch size, and $z_i$ and $z_j^{\xi}$ are the latent vectors:
\begin{align}
z_i &= h(f_{\theta}(x_i)) \\
z_j^{\xi} &= h(f_{\xi}(x_j)).
\end{align}
After training, $f_{\theta}$ is retained as the obtained feature extractor while other parts are discarded. SSL captures richer features than supervised learning, enabling task-agnostic and transferable representations that alleviate task-level overfitting in CIL. By learning rich, transferable features through SSL and maintaining a fixed encoder during the incremental process, the feature drift and confusion in CIL can be alleviated.

Experimentally, we measure feature richness and discrimination using intrinsic dimensionality and cosine similarity matrix, respectively, with details provided in the following.
\begin{figure*}[t]
	\centering
	\includegraphics[width=0.83\linewidth]{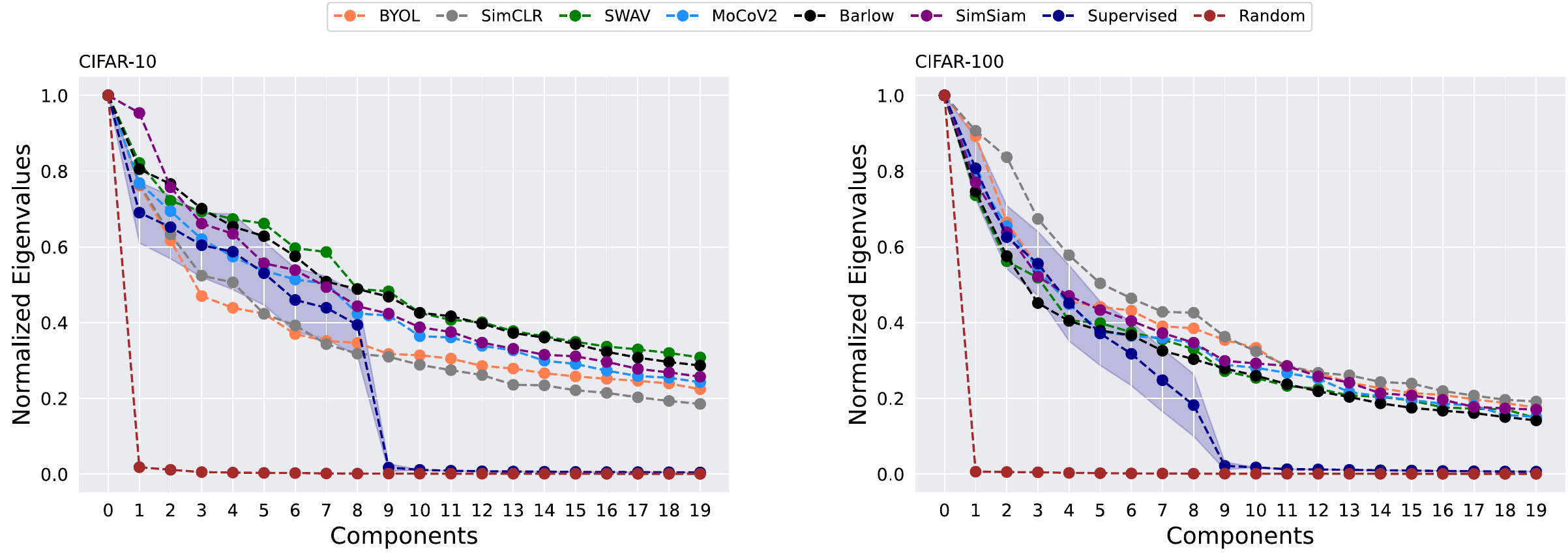}
	\vskip -0.1in
	\caption{The normalized singular values of the feature matrix, showing the curves of the first 20 normalized eigenvalues in descending order. Regardless of whether it is in-class (CIFAR10) or out-class (10 classes of CIFAR100), the normalized singular values decrease to 0 at N-1 for supervised learning and decrease to 0 at 1 for the randomly initialized network. The feature space of self-supervised learning has a higher PC-ID (as shown in Table\ref{table:0}) and its curve descends significantly slower.}
	\vskip -0.1in
	\label{fig:3}
\end{figure*}

\noindent\textbf{Intrinsic Dimensionality (ID)} is the minimal number of coordinates required to describe data representation without significant information loss \cite{ansuini2019intrinsic}. The ID of the feature space can be estimated using principal component analysis (PCA) on the normalized feature covariance matrix $\Sigma$. Mathematically, let $\lambda_i$ be the eigenvalues of $\Sigma$ sorted in descending order, then we can calculate the cumulative sum of the normalized eigenvalues as follows:
\begin{equation}
P(k) = \frac{\sum_{i=1}^{k} \lambda_i}{\sum_{i=1}^{N} \lambda_i},
\end{equation}
where $N$ is the total number of eigenvalues. The Intrinsic Dimensionality (PC-ID)~\cite{ansuini2019intrinsic} is defined as:

\begin{equation}
\text{PC-ID} = \min k \in \mathbb{N} : P(k) \geq 0.9 .
\end{equation}
Furthermore, the curve of the normalized eigenvalues, $\frac{\lambda_i}{\sum_{i=1}^{N} \lambda_i}$, provides a more nuanced depiction of the feature space. A larger-ID feature space will exhibit a curve that descends slower than a smaller-ID feature space \cite{yu2020learning,chen2021understanding}.

To compare the feature richness of self-supervised and supervised learning, we trained a ResNet18 network\cite{he2016deep} on the CIFAR10 dataset\cite{krizhevsky2009learning} using six self-supervised learning methods (BYOL\cite{grill2020bootstrap}, MoCoV2\cite{he2020momentum}, SimCLR\cite{chen2020simple}, SwAV\cite{caron2020unsupervised}, BarlowTwins\cite{zbontar2021barlow}, and SimSiam\cite{chen2021exploring} ) and supervised learning methods. Also included, a model with random initialization is included in comparison. After training, we performed PCA on the normalized feature covariance matrix to obtain the overall feature space of CIFAR10 and CIFAR100, respectively, for in-class and out-of-class evaluations. The PC-ID and the curves of normalized eigenvalues in descending order were then obtained.

 As illustrated in Figure \ref{fig:3} and Table \ref{table:0}, the PC-ID of the feature space obtained from self-supervised learning is significantly higher than that from supervised learning in both in-class (CIFAR10) and out-of-class (CIFAR100) evaluations. According to SSLR \cite{chen2021understanding}, supervised learning primarily aims to extract features that are relevant for closed-world classification, and its PC-ID is proportional to the number of classes, which is N-1 where N is the total number of classes. On the other hand, SSL focuses on learning class-agnostic features, which result in a much higher PC-ID and a more comprehensive feature space that is suitable for class incremental learning. The curves of normalized eigenvalues in descending order in Figure \ref{fig:3} demonstrate that the curve of SSL descends at a slower pace compared to SL, indicating that the feature space learned through SSL has a higher intrinsic dimensionality and the ability to represent a wider range of categories, thereby reducing representation confusion.

\begin{table}[htb]

 \small
 \centering 
 \renewcommand\tabcolsep{20pt}
 \renewcommand{\arraystretch}{1.3}
  \caption{PC-ID of the feature space generated by different learning methods. The models were trained on CIFAR10 using ResNet18 and tested on CIFAR10 and the data of  10 classes of CIFAR100 (disjoint classes with CIFAR10).}
\begin{tabular}{c|cc}

\toprule 
\hline
\multirow{2}{*}{\textbf{Method}}  & \multicolumn{2}{c}{\textbf{PC-ID}} \\ \cline{2-3} 
              & CIFAR10         & CIFAR100 \\ \hline
BYOL          &  240          &  177   \\
SimCL         &  210          &  188    \\
MoCoV2        &  175          &  155  \\
SWAV          &  235          &  184  \\
Barlow        &  263          &  191   \\
SimSiam       &  245          &  183   \\ 
Sup           &  9          &  9   \\  
Random        & 1           &  1     \\ \hline
  \bottomrule
\end{tabular}
 \label{table:0}

\end{table}

\begin{figure*}[t]
  \centering
  \includegraphics[width=1.0\linewidth]{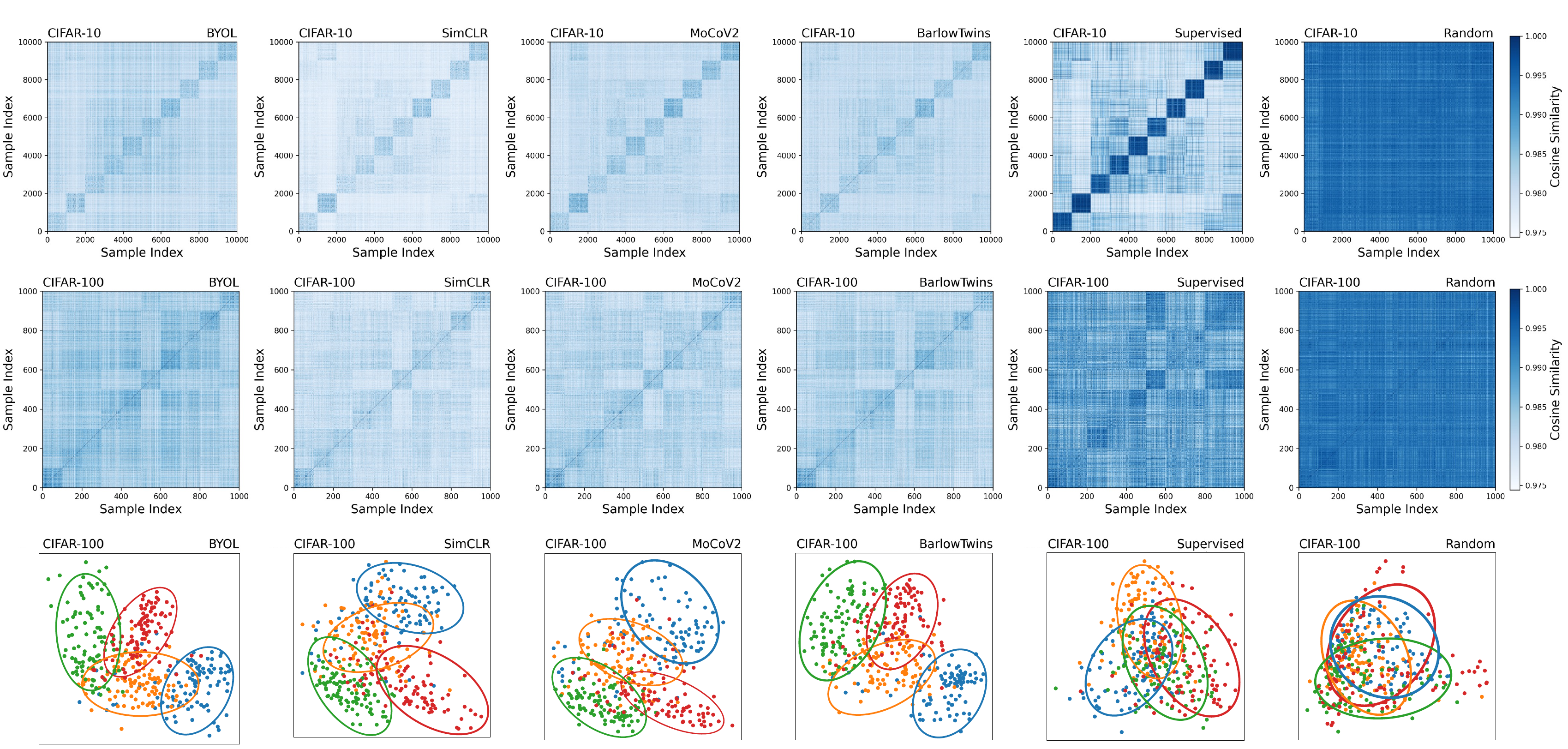}
  \caption{The Cosine similarity of the learned feature spaces through self-supervised learning and supervised learning evaluated in both in-class and out-of-class settings. The diagonal elements of the Cosine similarity matrix indicate the similarity between features of the same category, while the other elements represent the similarity between features of different categories. Using PCA to visualize the data of four categories in CIFAR100, it can be seen that the features extracted by supervised learning result in a more significant feature confusion when the categories are different from the training data.}
  \vskip -0.1in
  \label{fig:4}
  \vskip -0.1in
\end{figure*}

\textbf{Cosine similarity matrix} provides a visual representation of the similarity between different classes of features, and can directly reveal representation confusion. It calculates the cosine similarity between samples in the feature space, arranging the results by category. This means that the cosine similarity between samples of the same class is on the diagonal, while the other positions correspond to different classes. To examine representation confusion in CIL, we calculated the cosine similarity matrices for in-class and out-of-class evaluations. The encoder was trained on CIFAR10 with the same settings as previously described, and the cosine similarity matrices were calculated for both CIFAR10 and subsets of CIFAR100.

As depicted in Figure \ref{fig:4}, the in-class cosine similarity matrix demonstrates that the features of different classes learned through self-supervised learning are more orthogonal compared to supervised learning, meaning that the similarity between different classes is closer to zero. A feature space that is close to orthogonal between different classes provides better representation for new classes to emerge, thus alleviating representation confusion during CIL. The out-of-class cosine similarity matrix is used to evaluate the generalization performance of the feature representation across classes by using data from classes that are different from the training set. The PCA dimensionality visualization of the out-of-class feature distribution, along with the cosine similarity matrix, clearly shows that supervised learning leads to significant confusion in the feature representation compared to self-supervised learning.

\subsection{Incremeantal prototype classifier}
\label{subsection: prototype classifier}
The incremeantal prototype classifier (IPC) used is designed to alleviate classifier distortion during CIL, by simultaneously considering the discriminability of the current category, the generalizability of the new future category, and the anti-forgetting of the old category.
These factors are realized by \textbf{discriminative loss}, \textbf{regularization loss}, and \textbf{incremental update strategy}, respectively.

In our framework, The prototypes are denoted as $\varphi=\{\varphi_i|i=1,2,...,C\}$, where $\varphi_i\in \mathbb{R}^d$ and $d$ is the dimensionality of feature vectors. Given an input $x$, we first get its representation $f_{\theta}(x)$ by the pre-trained feature extractor, then we compare the feature with all prototypes and calculate the Euclidean distance between $f_{\theta}(x)$ and $\varphi_i$, \textit{i.e.}, $d_i(x) = ||f_{\theta}(x)-\varphi_i||_2^2$. 

\textbf{Discriminative loss.} IPC measures the similarity of samples and prototypes in the feature space by Euclidean distance. The probability that a sample $(x,y)$ matches the prototype $\varphi_i$, \textit{i.e.}, the sample belongs to class $i$, is calculated by distance based softmax:
\begin{equation}
  p(x \in \varphi_i|x)=\frac{e^{-\gamma d_i(x)}}{\sum_{k=1}^C e^{-\gamma d_k(x)}},
  \label{eq:1}
\end{equation}
where $\gamma$ is temperature scalar that affects the sparseness of the class distribution. Based on the probability defined, the cross entropy is used as discriminative loss:
\begin{align}
  l_{CE}\big((x,y);\varphi\big) = -\log p(x \in \varphi_y|x).
\end{align}

\textbf{Regularization loss.}
Minimizing the discriminative loss normally drives the prototype of correct class closer to the sample and the other prototypes away from the sample so as to improve the classification accuracy. Without regularization, however, loss minimization may lead to over-adjustment that makes the prototype of each class deviate from the sample distribution. The over deviation of prototype from sample distribution may deteriorate the generalization performance \cite{yang2020convolutional}. For CIL, the prototype deviation is even more influential to the performance of old classes. 

Inspired by DLQDF\cite{liu2004discriminative} and CPN\cite{yang2020convolutional}, which consider both discriminability and robustness, we introduce prototype learning loss (PL) to CIL, acting like a generative model to improve the generalization of the classifier in incremental learning.
Intuitively, PL brings the prototypes and corresponding samples closer together and reduces their confusion with new classes added during CIL. PL is also akin to maximum likelihood regularization of the Gaussian distribution, which is widely used in pattern recognition [11, 19]. PL is defined as follows: 
\begin{align}
\begin{split}
  l_{PL}\big((x,y);\varphi\big)=||f_{\theta}(x)-\varphi_y||_2^2.
\end{split}
\end{align}

With both discriminative and regularization loss functions, the final hybrid loss is defined as:
\begin{align}
  \begin{split}
  \mathcal{L} = l_{CE}+\lambda\cdot l_{PL},
  \label{eq:5} 
\end{split}
\end{align}
where $\lambda$ is a hyper-parameter.

\noindent\textbf{Incremental updating.} 
The prototype classifier learns a prototype in the feature space for the corresponding category by combining discriminative loss and regularization loss. In CIL, as new categories continue to emerge and old samples cannot be preserved (or only a small fraction of old samples can be preserved), the addition of new prototypes causes unreasonable shifts of the decision surface, i.e., classifier distortion. To alleviate classifier confusion, we design the updating rule as: while learning the new category prototype, the old prototypes participate in forward propagation and compute the loss together, but maintain their positions fixed by stopping the backward propagation. On one hand, considering the old prototypes jointly in calculating the discriminative loss guarantees discrimination between all classes.
On the other hand, gradient cutoff keeps the position of the old prototypes unchanged and alleviates catastrophic forgetting during CIL. Algorithm \ref{Algorithm: 1} illustrates the training and incremental update steps of IPC.


\begin{algorithm}[t]
\renewcommand{\algorithmicrequire}{\textbf{Input:}} 
\renewcommand{\algorithmicensure}{\textbf{Output:}} 
\caption{Training of Prototype Classifier in CIL}\label{Algorithm: 1}
\begin{algorithmic}
\REQUIRE training dataset $\mathcal{D}_t$ ,feature extractor $f_\theta(\cdot)$, old prototypes $\varphi_{old}$, batch size $N$
\ENSURE $\varphi = [\varphi_{old}, \varphi_{new}]$
\FOR{sampled minibatch $\{(x_k, y_k)_{k=1}^{N}\}$}
\FOR{all $k\in{1,2,...,N}$}
\STATE $z_k = \textbf{\textit{StopGrad}}(f_{\theta}(x))$ \hfill // Fix $f_\theta(\cdot)$
\STATE $\varphi = [\textbf{\textit{StopGrad}}(\varphi_{old}), \varphi_{new}]$ \hfill // Fix $\varphi_{old}$
\STATE $d_i(z_k) = ||z_k-\varphi_{i}||_2^2$, for all $\varphi_i\in\varphi$
\STATE Compute hybrid loss $\mathcal{L}_k$ with Eq. (\ref{eq:1})-(\ref{eq:5})
\ENDFOR
\STATE $\mathcal{L}_{batch}=\frac{1}{N}\sum_{k=1}^{N}\mathcal{L}_k$
\STATE Update $\varphi$ to minimize $\mathcal{L}_{batch}$ by backpropagation
\ENDFOR
\end{algorithmic}
\end{algorithm}

\noindent\textbf{Inference.} In the inference stage of IPC, the input $x$ is classified into the category to which its nearest prototype belongs. This is achieved by first computing the feature representation of the input $x$ using the trained encoder, and then computing the Euclidean distance between the feature representation and each prototype:
\begin{equation}
  x \in class\quad \arg\min_{i=1}^C d_i(x).
\end{equation}

This method of classification is simple, efficient, and has been shown to be effective in previous works, such as DLQDF and CPN\cite{yang2020convolutional}. Additionally, IPC can be easily combined with various self-supervised learning methods to further improve performance.

\subsection{A closer look at prototype classifier}
\label{subsec: closer look}

\begin{figure*}[t]
  \centering
  \includegraphics[width=0.9\linewidth]{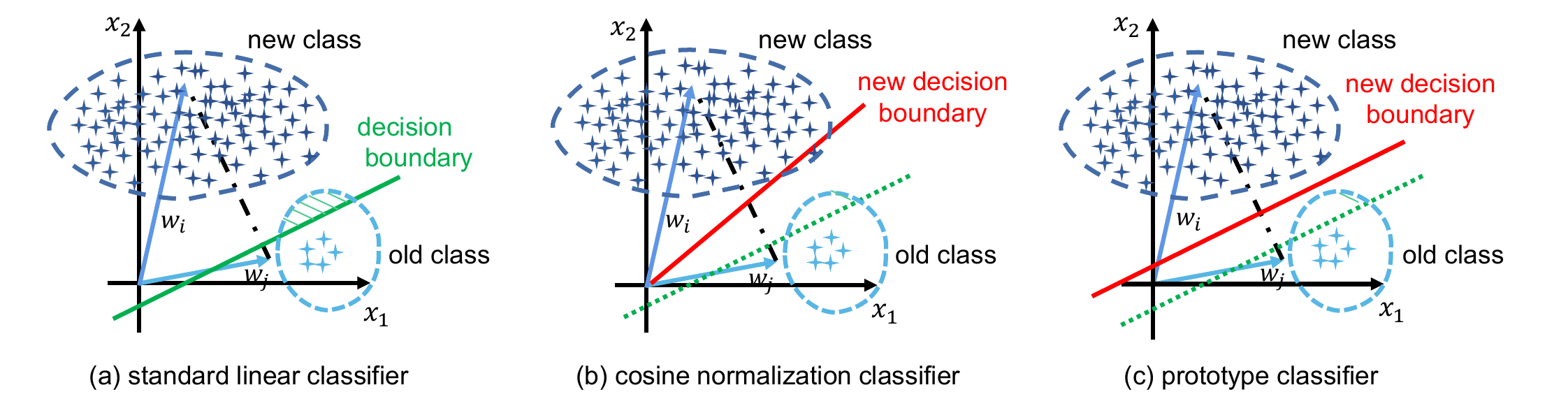}
  \vskip -0.1in
  \caption{Comparison of standard linear classifier, cosine normalization classifier\cite{hou2019learning} and prototype classifier. The standard linear classifier may misclassify some samples due to the unbalanced dataset. The cosine normalization classifier changes the decision boundary to angle bisector of the weight vector to alleviate this problem, but it would be less effective when the data is unevenly distributed along the angle. In CIL  scenarios, the prototype classifier, designed to produce compact decision regions as weight values increase, obtains optimal classification boundaries by discriminative learning.}
   \vskip -0.1in
  \label{fig:5}
\end{figure*}

\begin{figure*}[t]
  \centering
  \includegraphics[width=1.0\linewidth]{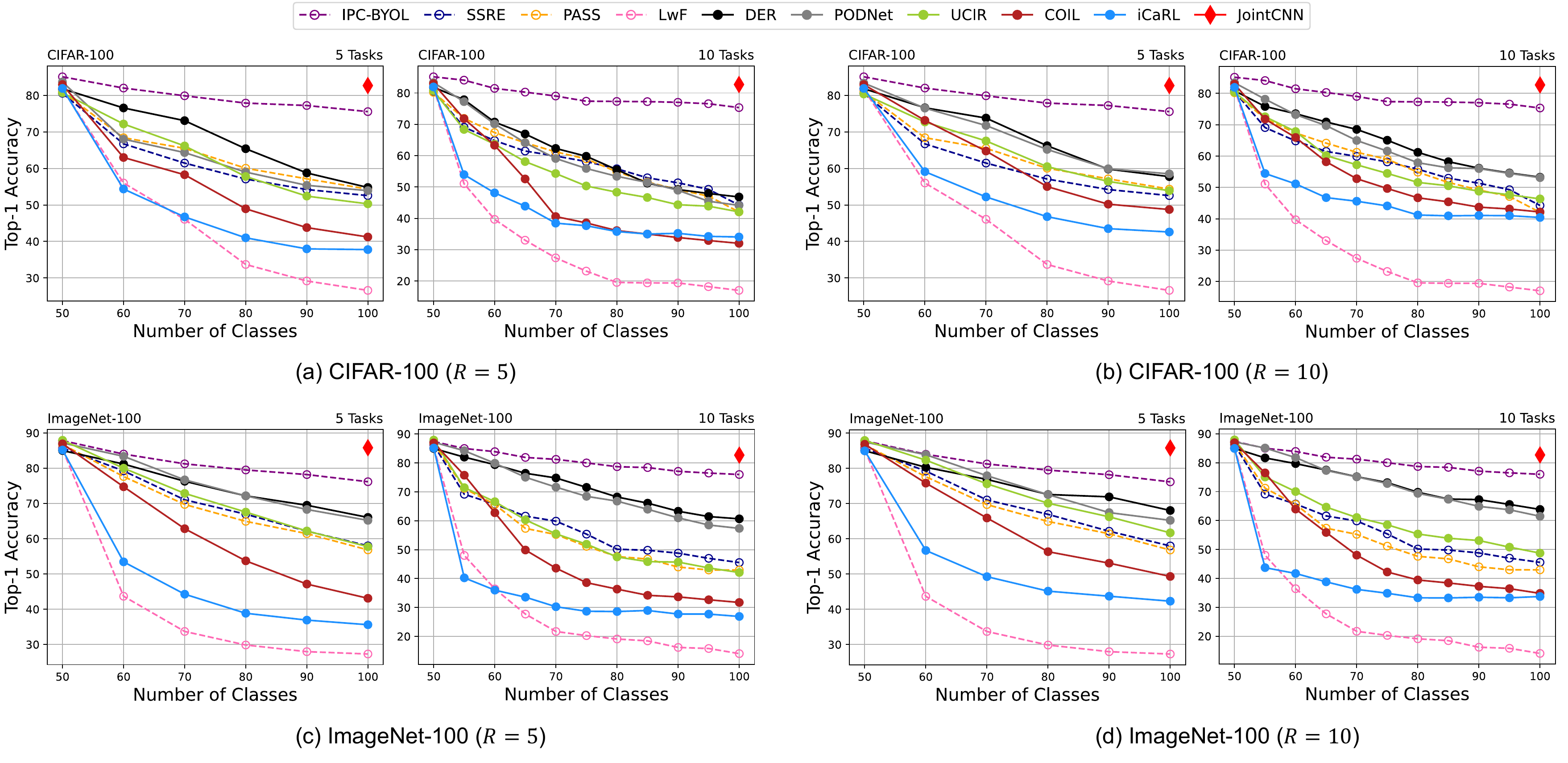}
  \vskip -0.1in
  \caption{The accuracy curves for different methods across each task in 5 and 10 tasks settings on CIFAR-100 (a-b) and ImageNet-100 (c-d) when $R=5$ and $R=10$. Our method exhibits significantly higher accuracy in CIL without requiring the preservation of samples from previous tasks.}
   \vskip -0.1in
  \label{fig:6}
\end{figure*}

\begin{figure*}[t]
  \centering
  \includegraphics[width=1.0\linewidth]{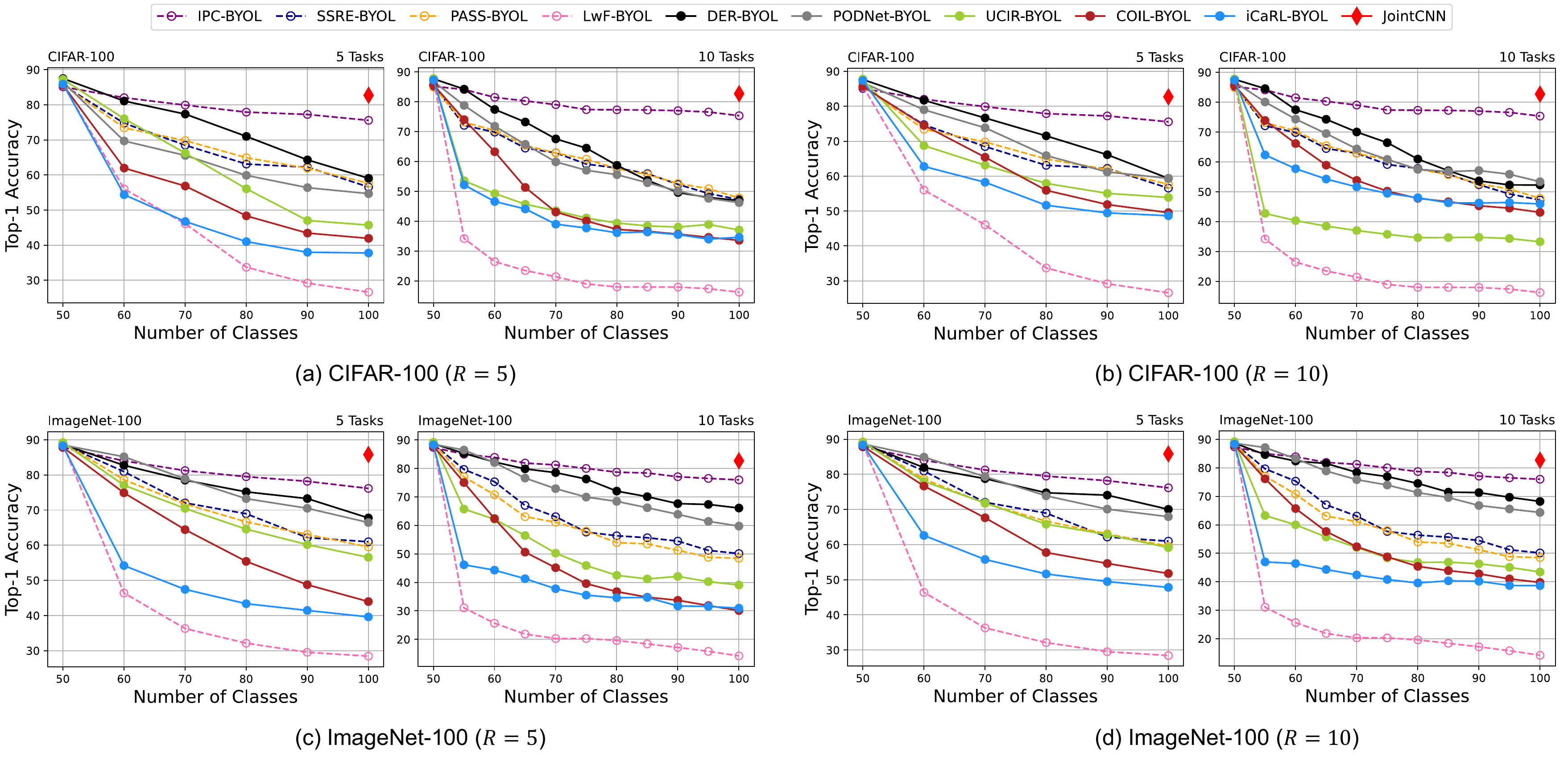}
  \vskip -0.1in
	\caption{Using the pre-trained BYOL model, accuracy curves for different methods across tasks in 5 and 10 task settings on CIFAR-100 (a-b) and ImageNet-100 (c-d) with $R=5$ and $R=10$. Our method outperforms others in CIL when also using pre-trained models.}
   \vskip -0.1in
  \label{fig:7}
\end{figure*}

Here we make a more in-depth analysis of the prototype classifier to explain why our method is suitable for CIL. 
For classification, the prototype classifier's discriminant function for class $i$ is as follow:
\begin{align}
\begin{split}
  g_i(x) &= -||f_{\theta}(x)-\varphi_{i}||_2^2\\
  &= -(f_{\theta}(x)-\varphi_{i})^T (f_{\theta}(x)-\varphi_{i})\\
  &= 2\varphi_{i}^T f_{\theta}(x)- \varphi_{i}^T\varphi_{i} - f_{\theta}(x)^T f_{\theta}(x).
  \label{eq:8}
\end{split}
\end{align}
It can be deduced that the decision boundary is linear, as $g_i(x)=g_j(x)$ leads to:
\begin{equation}
  (\varphi_{i}^T-\varphi_{j}^T) f_{\theta}(x)- (\varphi_{i}^T\varphi_{i}-\varphi_{j}^T\varphi_{j})/2 = 0.
  \label{eq:10}
\end{equation}
While we can get the decision boundary of standard linear classifier:
\begin{equation}
  (\mathbf{\omega}_{i}^T-\mathbf{\omega}_{j}^T) f_{\theta}(x) + (b_i-b_j)= 0,
  \label{eq:11}
\end{equation}
where $\varphi$ and $\mathbf{\omega},b$ are learnable parameters estimated by minimizing discriminative loss.
A key problem that limits the performance of CIL is the significant imbalance between the old classes seen at previous phases and the new ones at the current phase. With the standard linear classifier, the magnitudes of both $\mathbf{\omega}$ and $b$ for the new classes will be significantly higher than those for the old classes. And this results in the predictions that favor new classes\cite{hou2019learning}. The above phenomenon inspired Hou \textit{et al.} \cite{hou2019learning} to propose the \textit{cosine normalization} classifier to rebalance the decision boundary of the standard linear classifier. They accomplish this goal by changing the decision boundary to angle bisector of the weight vector, but it will lose its effect when the data is unevenly distributed along the angle. On the other hand, our prototype classifier can overcome data imbalance in all cases of linear separation.

%

By examining Equations (\ref{eq:10}) and (\ref{eq:11}), we can deduce that the difference between prototype classifiers and linear classifiers lies in the bias term of the prototype classifier, which is the Euclidean norm of the weights. Unlike linear classifiers, prototype classifiers exhibit smaller decision regions for larger weights, effectively suppressing the tendency to learn substantially larger weights for new classes during class incremental learning. Intuitively, since the weights of the prototype classifier are directly transformed from the class prototypes, they are inherently less sensitive to class imbalance. The comparison of the three classifiers in terms of their decision boundaries is shown in Figure \ref{fig:5}.

\begin{table*}[htb]
\small
 \centering 
 \renewcommand\tabcolsep{14pt}
 \renewcommand{\arraystretch}{1.3}	
  \caption{Comparisons of average incremental accuracy on CIFAR-100 and ImageNet-100 under 5/10-step-5/10-replay settings between state-of-the-art methods and our method ($R$=0). The methods irrelevant of R are non-exemplar based methods.}
\begin{tabular}{c|cccc|cccc}
\toprule 
\hline
\multirow{2}{*}{\textbf{Method}}                                                         & \multicolumn{4}{c|}{\multirow{2}{*}{\textbf{CIFAR-100}}}            & \multicolumn{4}{c}{\multirow{2}{*}{\textbf{ImageNet-100}}}                  \\
                                                                                         & \multicolumn{4}{c|}{}                                               & \multicolumn{4}{c}{}                                                        \\ \hline
\multirow{2}{*}{\begin{tabular}[c]{@{}c@{}}Average accurary\\ w/o pretrian\end{tabular}} & \multicolumn{2}{c|}{T=5}               & \multicolumn{2}{c|}{T=10}  & \multicolumn{2}{c|}{T=5}               & \multicolumn{2}{c}{T=10}           \\ \cline{2-9} 
                                                                                         & R=5     & \multicolumn{1}{c|}{R=10}    & R=5          & R=10        & R=5     & \multicolumn{1}{c|}{R=10}    & R=5               & R=10           \\ \hline
DER                                                                                      & 68.41   & \multicolumn{1}{c|}{69.34}   & 60.92        & 65.27       & 75.47   & \multicolumn{1}{c|}{75.78}   & 71.71             & 73.28          \\
PODNet                                                                                   & 64.06   & \multicolumn{1}{c|}{69.21}   & 59.38        & 64.42       & 75.02   & \multicolumn{1}{c|}{75.74}   & 70.36             & 73.27          \\
UCIR                                                                                     & 63.26   & \multicolumn{1}{c|}{65.24}   & 54.61        & 57.95       & 71.37   & \multicolumn{1}{c|}{73.98}   & 56.22             & 61.73          \\
COIL                                                                                     & 56.37   & \multicolumn{1}{c|}{62.48}   & 47.24        & 54.75       & 61.40   & \multicolumn{1}{c|}{64.55}   & 47.83             & 50.89          \\
iCaRL                                                                                    & 49.96   & \multicolumn{1}{c|}{54.36}   & 43.48        & 48.04       & 49.03   & \multicolumn{1}{c|}{53.68}   & 35.80             & 40.65          \\
SSRE                                                                                     & \multicolumn{2}{c|}{62.11}             & \multicolumn{2}{c|}{58.87} & \multicolumn{2}{c|}{70.51}             & \multicolumn{2}{c}{58.11}          \\
PASS                                                                                     & \multicolumn{2}{c|}{64.41}             & \multicolumn{2}{c|}{58.95} & \multicolumn{2}{c|}{69.54}             & \multicolumn{2}{c}{55.60}          \\
LWF                                                                                      & \multicolumn{2}{c|}{45.55}             & \multicolumn{2}{c|}{31.78} & \multicolumn{2}{c|}{41.29}             & \multicolumn{2}{c}{29.36}          \\ \hline
\multirow{2}{*}{\begin{tabular}[c]{@{}c@{}}Average accurary\\ w/ pretrian\end{tabular}}  & \multicolumn{2}{c|}{T=5}               & \multicolumn{2}{c|}{T=10}  & \multicolumn{2}{c|}{T=5}               & \multicolumn{2}{c}{T=10}           \\ \cline{2-9} 
                                                                                         & R=5     & \multicolumn{1}{c|}{R=10}    & R=5          & R=10        & R=5     & \multicolumn{1}{c|}{R=10}    & R=5               & R=10           \\ \hline
DER-BYOL                                                                                 & 73.37   & \multicolumn{1}{c|}{73.84}   & 64.67        & 66.95       & 77.68   & \multicolumn{1}{c|}{78.04}   & 75.83             & 77.07          \\
PODNet-BYOL                                                                              & 65.43   & \multicolumn{1}{c|}{71.01}   & 61.13        & 65.12       & 77.13   & \multicolumn{1}{c|}{77.46}   & 72.39             & 75.02          \\
UCIR-BYOL                                                                                & 63.04   & \multicolumn{1}{c|}{64.42}   & 46.61        & 41.25       & 69.67   & \multicolumn{1}{c|}{71.14}   & 52.26             & 54.28          \\
COIL-BYOL                                                                                & 56.35   & \multicolumn{1}{c|}{63.81}   & 48.65        & 55.99       & 62.55   & \multicolumn{1}{c|}{66.08}   & 47.98             & 54.67          \\
iCaRL-BYOL                                                                               & 50.62   & \multicolumn{1}{c|}{59.70}   & 43.96        & 54.15       & 52.39   & \multicolumn{1}{c|}{59.28}   & 41.53             & 46.06          \\
SSRE-BYOL                                                                                & \multicolumn{2}{c|}{68.44}             & \multicolumn{2}{c|}{61.50} & \multicolumn{2}{c|}{72.29}             & \multicolumn{2}{c}{63.56}          \\
PASS-BYOL                                                                                & \multicolumn{2}{c|}{68.89}             & \multicolumn{2}{c|}{62.00} & \multicolumn{2}{c|}{71.47}             & \multicolumn{2}{c}{61.36}          \\
LWF-BYOL                                                                                 & \multicolumn{2}{c|}{46.21}             & \multicolumn{2}{c|}{27.21} & \multicolumn{2}{c|}{43.59}             & \multicolumn{2}{c}{26.64}          \\
\cdashline{1-9}
\rowcolor{mygray}\textbf{OUR}                                                                             & \multicolumn{2}{c|}{\textbf{79.62}}             & \multicolumn{2}{c|}{\textbf{79.16}} & \multicolumn{2}{c|}{\textbf{81.15}}    & \multicolumn{2}{c}{\textbf{80.55}} \\ \hline
\bottomrule
\end{tabular}

 \label{table:1}
 \vskip -0.1in
\end{table*}

\section{Experiments}
\label{sec: 5}
\subsection{Setup}
\noindent\textbf{Datasets.} 
We verify the effectiveness of our proposed method using two well-known benchmark datasets for CIL, namely CIFAR-100 \cite{krizhevsky2009learning} and ImageNet-100 \cite{deng2009imagenet}. The classes were sorted in a predetermined order and divided into consecutive subsets. Following the typical evaluation protocol\cite{hou2019learning,hu2021distilling,simon2021learning}, we first train our algorithm on half of the classes in each dataset, and then gradually add the remaining classes in subsequent phases.

\noindent\textbf{Comparison Methods.} We compare with state-of-the-art data replay-based approaches in the field of CIL, including iCaRL\cite{rebuffi2017icarl}, LUCIR\cite{hou2019learning}, and PODNet\cite{douillard2020podnet}, as well as data-free methods such as LwF\cite{li2017learning}, PASS\cite{zhu2021prototype}, and SSRE\cite{zhu2022self}. We implemented these methods using open-source code. The PyCIL\cite{zhou2021pycil} repository played a crucial role in organizing and supplying the code for many of the continual learning methods.

\noindent\textbf{Fair Comparison of Pre-trained Model Use.} To ensure a fair comparison, the pre-trained model  is used with all comparison methods in our experiments. All comparison methods utilized the ResNet50\cite{he2016deep} architecture, which is identical to the pre-trained model. To prevent prior exposure to the classes that would appear in the continual learning process, the classes displayed in the CIL stage were excluded from the pre-training stage. The classes from CIFAR-100\cite{krizhevsky2009learning} and ImageNet-100\cite{deng2009imagenet} were removed from the ImageNet\cite{deng2009imagenet} dataset to form the pre-training dataset.

\noindent\textbf{Implementation details.} In our experiments. The Stochastic Gradient Descent (SGD) optimizer\cite{krizhevsky2012imagenet} is employed to train the networks with an initial learning rate of 0.1 and momentum of 0.9. The learning rate decay during training is set through the cosine learning rate decay\cite{loshchilov2016sgdr}. All models were trained with a batch size of 128 and 160 epochs per phase, and the experiments were repeated three times with the average results reported. We tested different incremental settings, with $T$ phases (\textit{e.g.}, 5 and 10) for CIFAR-100 and ImageNet-100. For the exemplar-based methods, $R$ samples per old class were selected using the \textit{herd selection method}\cite{rebuffi2017icarl}. The CIL setting can be denoted as $T$-step-$R$-replay. \textit{Our code will be made public once this paper is accepted, and more implementation details can be found in the code.}

\subsection{Comparative results}
To conduct a fair comparison, we utilized the PyCIL library to replicate recent classical continual learning methods and used ResNet50 backbone to ensure fairness. For exemplar-based methods, comparison considered reserving 5-sample and 10-sample per old class settings and tested on 5 tasks or 10 tasks under the iCaRL setting.

The comparative results of various methods on the benchmark datasets are presented in Table \ref{table:1} and Figure \ref{fig:6} - \ref{fig:7}. it is shown our method achieves a higher average accuracy and outperforms other methods.

\noindent\textbf{Evaluation on CIFAR-100.}
As shown in Figure \ref{fig:6} (a) and (b), the dotted and solid lines represent exemplar-based methods and non-exemplar based methods, respectively. The red diamond marks the result of collecting all samples for training as the upper performance bound. (a) and (b) show the results of exemplar-based methods trained by preserving 5 and 10 samples per category. Among the previous methods, exemplar-based methods generally outperform non-exemplar methods. However, our non-exemplar based IPC method even outperforms the state-of-the-art (SOTA) exemplar-based methods, with a margin of 10.28\%-18.25\%, under the settings of 5 or 10 incremental tasks with 5 or 10 samples preserved per category.

\noindent\textbf{Evaluation on ImageNet-100}
As shown in Figure \ref{fig:6} (c) and (d), the results observed in ImageNet-100 match those in CIFAR-100. Previous works have shown that exemplar-based methods outperform non-exemplar based methods, especially in the ImageNet100 task. Despite this, our IPC method still surpasses the SOTA exemplar-based methods, even when they store 5 or 10 samples, showcasing its effectiveness. When comparing to exemplar-based methods with the same settings of 5 or 10 incremental tasks and 5 or 10 stored samples, IPC demonstrates an improvement of 5.37\%-8.44\%.

\noindent\textbf{Fair Evaluation with Pre-training.} 
For fair comparison, we added self-supervised pre-trained models to the comparison methods and the pre-trained models were trained on task-agnostic data. Results shown in Figure \ref{fig:7} indicate that the pre-trained model improved the performance of the comparison method initially, but failed to prevent catastrophic forgetting as the number of tasks increased.
CIL is a series of fine-tuning processes, and supervised learning can lead to degradation of the encoder, as discussed in Sec \ref{subsec: Learning representations via SSL}. Overfitting to the task can cause the encoder to only focus on features useful for classification, leading to representation confusion when new categories are introduced.

Our method prevents encoder degradation by fixing it and utilizing a prototype classifier with an incremental update strategy to alleviate classifier distortion, resulting in improved performance when using pre-trained models for continual learning tasks. The experimental results on CIFAR-100 and ImageNet-100 datasets show that our method outperforms the SOTA methods when they save a maximum of 5 or 10 samples per class and perform 5 or 10 incremental tasks, with an improvement of 5.78\%-14.49\% and 3.11\%-4.73\% respectively.

\subsection{Ablation study}
To evaluate the effects on representation drift, confusion, and classifier distortion in incremental learning, our model consists of two components: a self-supervised pre-trained encoder and an incrementally updated classifier (IPC). We analyze the impact of each component on the performance. The encoder can be implemented by various pre-training methods, including supervised and self-supervised methods, and our model is compatible with these different pre-training methods. For the classifier part, we evaluate different types of classifiers in this model framework, including linear classifiers, cosine linear classifiers, and NME classifiers. Then, within the IPC we investigate the effects of different learning options, including the PL loss with generative attributes, and the incremental update strategy (IUS). Finally, we also investigate the effect of the only one hyperparameter, the lambda of the PL loss, by varying its values. All of these experiments were conducted on the CIFAR-100 dataset, and when exploring the classifiers, we used a BYOL pre-trained encoder, as the choice of pre-training method does not affect the conclusion of this part.

\begin{figure}[h]
  \centering
  \includegraphics[width=1.0\linewidth]{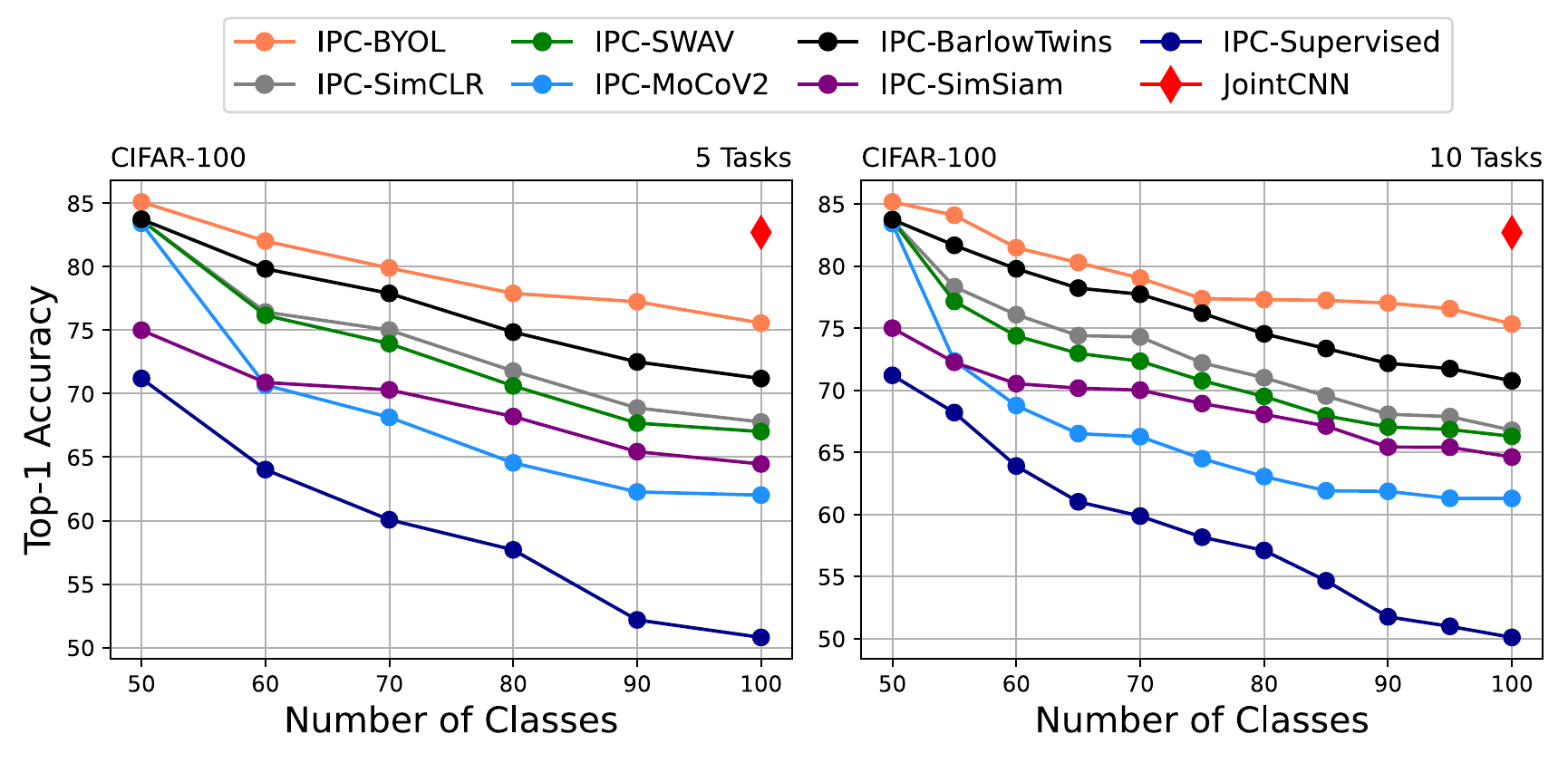}
  \vskip -0.1in
  \caption{The accuracy curves for our approach with 5 and 10 task settings on CIFAR-100, when using different pre-trained models including self-supervised and supervised models. The results indicate that self-supervised methods yield superior performance.}
\label{fig:8}
\end{figure}

\begin{figure}[h]
  \centering
  \includegraphics[width=1.0\linewidth]{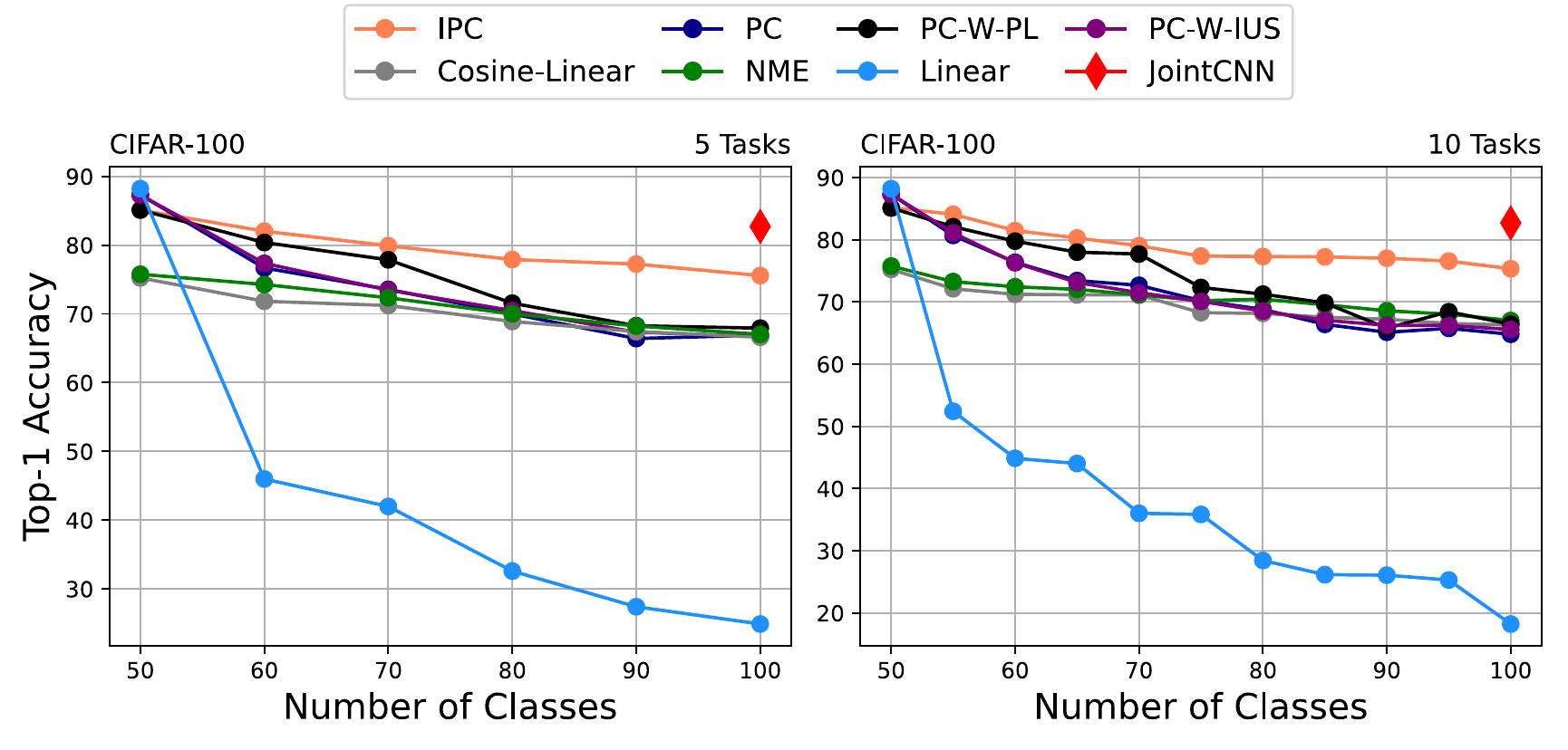}
  \vskip -0.1in
  \caption{The accuracy of our method, when employing different classifiers, including IPC classifiers w/o prototype learning loss (PL loss) and w/o incremental update strategy(IUS), is evaluated in 5-task and 10-task settings on CIFAR-100.}
  \vskip -0.1in
  \label{fig:9}
\end{figure}

\begin{table}[htb]

 \small \centering 
 \renewcommand\tabcolsep{7pt}
 \renewcommand{\arraystretch}{1.3}
   \caption{The average accuracy of our approach, employing a variety of pre-trained models such as self-supervised and supervised models, assessed in both 5-task and 10-task on CIFAR-100.}
\begin{tabular}{c|c|cc}

\toprule 
\hline
\multirow{2}{*}{\textbf{Method}} & \multirow{2}{*}{\textbf{Self-Supervised}} & \multicolumn{2}{c}{\textbf{Average Accurary(\%)}} \\ \cline{3-4} 
                   &               & 5 Tasks         & 10 Tasks \\ \hline
IPC-MoCoV2         & \Checkmark    & 68.51          &   66.47   \\
IPC-SimCLR         & \Checkmark    & 73.92          &  72.93    \\
IPC-SWAV           & \Checkmark    & 73.17          &   71.71  \\
IPC-Barlow         & \Checkmark    & 76.66          &   76.35  \\
PC-SimSiam         & \Checkmark    & 69.04          &  68.86   \\
IPC-Sup            & \XSolidBrush  & 59.32          &   58.81   \\ \cdashline{1-4} 
\textbf{IPC-BYOL}  & \Checkmark    & \textbf{79.62} & \textbf{79.16}     \\ \hline
  \bottomrule
\end{tabular}
 \label{table:2}
  \vskip -0.1in
\end{table}

\noindent\textbf{Impact of Different Pre-trained Encoders.} 
To investigate the effects of different pre-training models, we tested various self-supervised pre-training methods including BYOL\cite{grill2020bootstrap}, MoCoV2\cite{he2020momentum}, SimCLR\cite{chen2020simple}, SwAV\cite{caron2020unsupervised}, BarlowTwins\cite{zbontar2021barlow}, and SimSiam\cite{chen2021exploring} within our framework. For comparison, we also tested supervised pre-training methods, which were trained on the ImageNet dataset excluding the CIFAR-100 related categories.
As analyzed in Section \ref{sec: 2}, self-supervised methods learned more general and category-agnostic features, leading to better performance in our framework compared to the supervised methods as shown in Table \ref{table:2} and Figure \ref{fig:8}, with an increase of 9.19\%-20.3\% and 7.66\%-20.35\% respectively in 5 and 10 steps. Among these self-supervised methods, BYOL was found to be the most suitable, possibly due to its similarity measurement method based on the Euclidean distance.


\begin{table}[htb]

 \small \centering 
 \renewcommand\tabcolsep{4pt}
 \renewcommand{\arraystretch}{1.3}
   \caption{The average accuracy of our approach, using different classifiers and assessing in both 5-task and 10-task settings on the CIFAR-100.}
\begin{tabular}{c|cc|cc}

\toprule 
\hline
\multirow{2}{*}{\textbf{Classifier}} & \multirow{2}{*}{\textbf{PL Loss}} & \multirow{2}{*}{\textbf{StopGrad}} & \multicolumn{2}{c}{\textbf{Average Accurary(\%)}} \\ \cline{4-5} 
                            &              &              & 5 Tasks   & 10 Tasks \\ \hline
Linear                      & \XSolidBrush &\XSolidBrush  &  43.50    &  38.68   \\
Cosine Linear               & \XSolidBrush &\XSolidBrush  &  70.20    &  69.53    \\
NME                         & \XSolidBrush &\XSolidBrush  &  71.27    &  70.76  \\
Prototype                   & \XSolidBrush &\XSolidBrush  &  73.47    &  71.95  \\
PC+PL                       & \Checkmark   & \XSolidBrush &  75.18    &  74.24   \\
PC+IUS                      & \XSolidBrush & \Checkmark   &  73.77    &  72.09   \\ 
  \cdashline{1-5}
\textbf{PC+PL+IUS}          & \Checkmark   &\Checkmark    &  \textbf{79.62}      & \textbf{79.16}     \\ \hline
  \bottomrule
\end{tabular}
 \label{table:3}
\end{table}
\noindent\textbf{Impact of Different Classifiers.} In this ablation study, we examine the effects of various classifiers on our framework, including the standard linear classifier, cosine normalization classifier, NME classifier, and our proposed incremental prototype classifier (IPC).

During the incremental learning process, linear classifiers tend to assign larger weights to new classes in the absence of old samples. This leads to misclassifying old classes as new ones, causing significant classifier distortion. As shown in Figure \ref{fig:9} and Table \ref{table:3}, the cosine linear classifier and NME classifier address this issue through weight normalization or weights from sample distribution. Although this alleviates forgetting, as discussed in Section \ref{subsec: closer look}, it still adversely impacts classification accuracy.

Our IPC is a linear classifier learning prototypes discriminatively while maintaining the compactness of samples per class. Utilizing backpropagation and incremental update strategies, it effectively identifies appropriate decision boundaries, considerably alleviating classifier distortion.


\noindent\textbf{Impact of PL Loss and Incremental Update Strategy.} We separately test the effects of PL loss abd incremental update strategy (IUS) on IPC. The PL loss allows the model to guide the prototypes to represent class sample distribution, making the classifier more robust in open-set environments. The IUS employs gradient clipping, enabling IPC to learn prototypes of new classes while maintaining the positions of the old classes. The results in Table \ref{table:3} show that The combination of these two components plays a crucial role in alleviating classifier distortion.

\begin{figure}[t]
  \centering
  \includegraphics[width=1.0\linewidth]{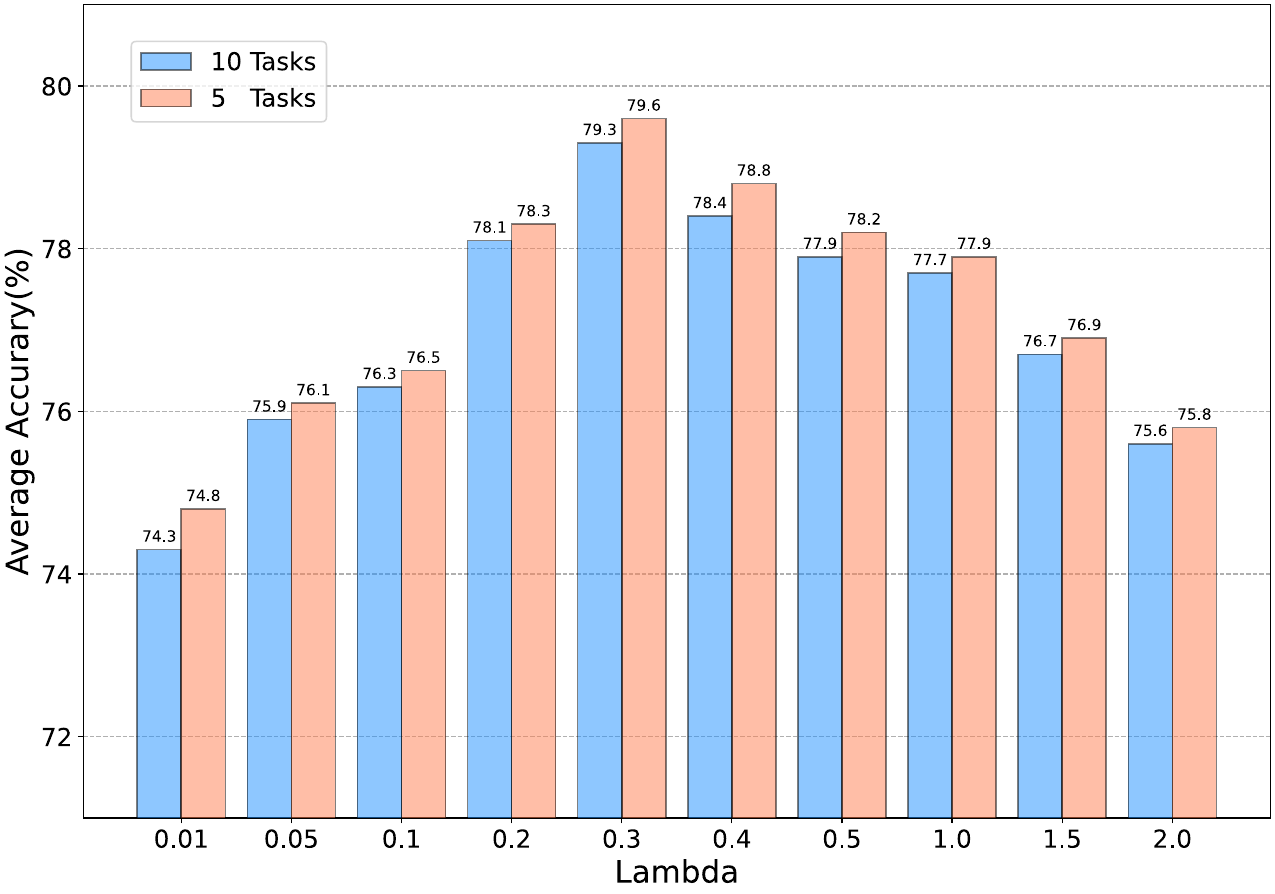}
  \vskip -0.1in
  \caption{The influence of varying lambda values, which is the only hyperparameter in our method, on the average accuracy in CIFAR-100 is examined for both 5-task and 10-task settings.}
  \vskip -0.1in
  \label{fig:10}
\end{figure}

\noindent\textbf{Impact of Hyperparameter.} In our framework, the only hyperparameter is the coefficient lambda for the PL loss. It is to balance the weights of generative loss and discriminative loss, and is needed to scale the PL loss to a reasonable range. As shown in Figure \ref{fig:10}, excessively high or low coefficients are detrimental to the average accuracy. A value around 0.3 gives optimal performance, while values slightly deviating from it do not deteriorate the performance sharply.

\begin{figure}[t]
  \centering
  \includegraphics[width=1.0\linewidth]{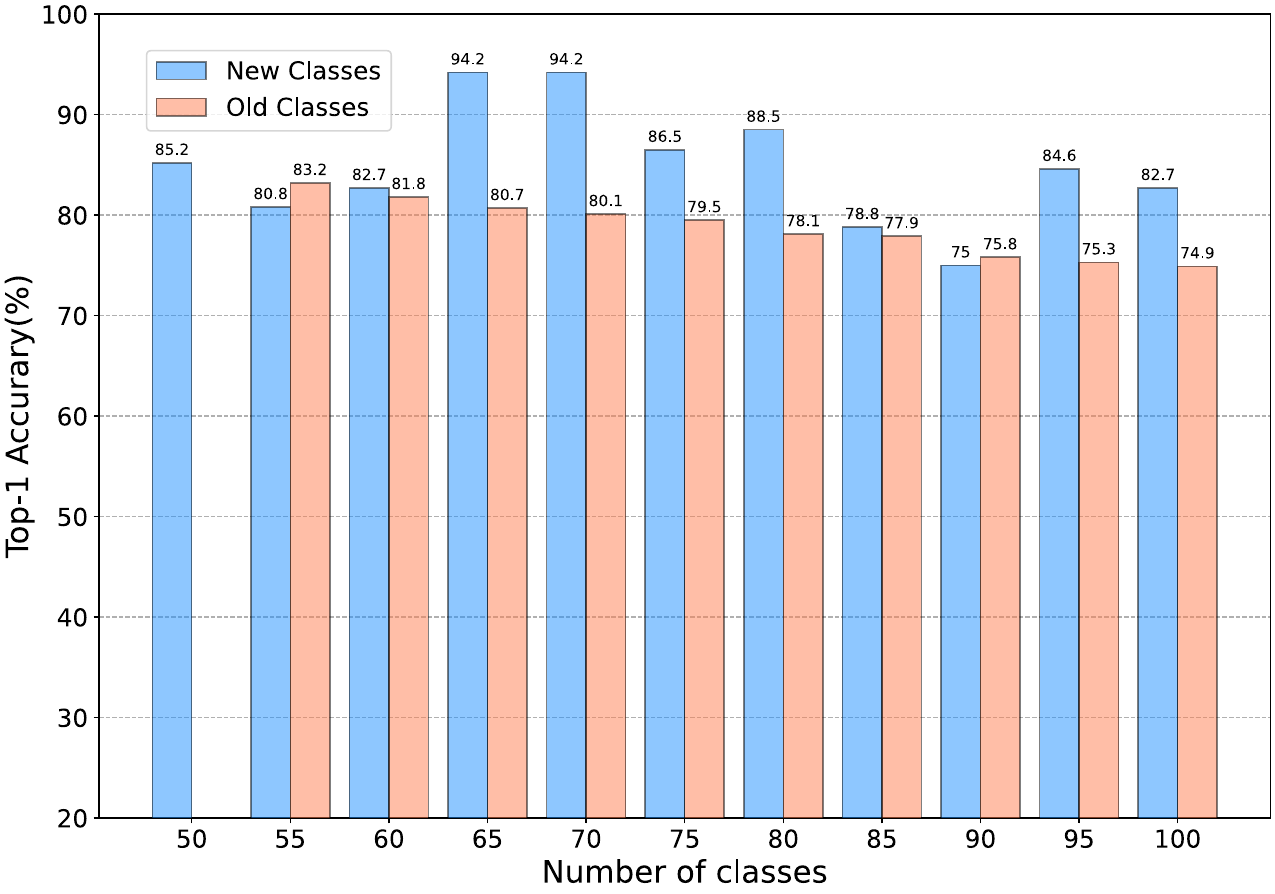}
  \vskip -0.1in
  \caption{The accuracy of our method on new and old classes in the CIFAR-100 dataset under a 10-task incremental learning setting.}
  \vskip -0.1in
  \label{fig:11}
\end{figure}
\noindent\textbf{Accuracy Across New and Old Classes.} In our ablation study, we evaluated the performance of our method on both new and old classes on the CIFAR-100 dataset under a 10-task incremental learning setting. We concentrated on the balance of accuracy, highlighting our method's effectiveness in learning new information while retaining knowledge of old classes. As depicted in Figure \ref{fig:11}, our method displays rather balanced accuracies across new and old categories. This indicates that our method successfully mitigates catastrophic forgetting and enhances overall performance in incremental learning tasks.
\section{Conclusion}
\label{sec: 6}
To tackle catastrophic forgetting in CIL, which is primarily caused by three factors - \textbf{\textit{representation drift}}, \textbf{\textit{representation confusion}}, and \textbf{\textit{classifier distortion}} - we introduce a simple yet effective two-stage learning framework with a fixed encoder and an incrementally updated classifier.
The encoder, trained through SSL, attains a feature space with high intrinsic dimensionality. By keeping the encoder fixed, we effectively address representation confusion and drift. Meanwhile, the IPC classifier leverages prototype learning and an incremental update strategy to minimize classifier distortion.
Remarkably, this innovative framework improves the performance remarkably even without reserved samples, encouraging CIL researchers to pay more attention to SSL for representation learning for CIL.

Our approach can be extended in several ways in the future. First, the SSL+IPC framework can be allied to exemplar-based CIL by slightly adjusting the prototype learning procedure. Second, for CIL for long range of tasks, the adaptation of feature representation may benefit. So, the pre-trained representation model needs to be updated in incremental SSL. Recent continuous self-supervised learning methods, such as CaSSLe\cite{fini2022self} and Continual BarlowTwins\cite{marsocci2022continual}, can offer valuable insights in this regard.
Third, the SSL+CIL framework can be combined with various CIL strategies, and under this framework, further improvements of SSL methods, classifiers, incremental learning strategies can be considered.

%
%
%


\bibliography{refer}
\bibliographystyle{IEEEtran}
 
%

\newpage

%
%
%
%
%
%
%
%
\vfill

\end{document}